\documentclass[letterpaper]{article}
\usepackage{graphicx}
\usepackage{aaai}
\usepackage{times}
\usepackage{helvet}
\usepackage{courier}
\usepackage{hyperref}
\nocopyright
 
\begin{document}
%
\title{PragAlign: Feedback-Guided Pragmatic Alignment for Controlled Synthetic Dialogue Generation}
\author{Smitha Muthya Sudheendra, Jaideep Srivastava \\
University of Minnesota, Twin Cities \\
muthy009@umn.edu, srivasta@umn.edu
}

\maketitle
\begin{abstract}
\begin{quote}
Synthetic dialogue generation can support research in privacy-restricted service settings, but generated conversations must preserve communicative intent, affective meaning, and natural dialogue flow. We introduce PragAlign, a feedback-guided framework for controlled synthetic dialogue generation conditioned on service context, target intent, and target emotion, with auxiliary trait-style controls. PragAlign uses a generate--evaluate--revise loop in which an LLM-based evaluator scores intent alignment, emotion alignment, coherence, fluency, and aggregate quality, then provides criterion-specific feedback for up to three refinement rounds. On 800 matched dialogue specifications, PragAlign achieves 99.50\% evaluator-defined acceptance, compared with 72.25\% for one-shot generation and 95.88\% for repeated generation without structured feedback. This indicates that repeated attempts account for much of the gain over one-shot generation, while structured feedback primarily improves last-mile multi-constraint satisfaction rather than broad average quality. Refinement gains are concentrated in emotion alignment, which is also the dominant failure mode in ablations. A separate human evaluation of 1,200 generated dialogues shows that intent expression and dialogue flow are highly recognizable to annotators, while emotion appropriateness is less stable and more subjective. These results support PragAlign as a quality-control framework for improving evaluator-defined communicative constraint satisfaction, while showing that affective realization and independent human-perceived quality remain open challenges.
\end{quote}
\end{abstract}

\section{Introduction}
\label{sec:introduction}

Synthetic dialogue data can support the development and evaluation of conversational agents, especially in service domains where real customer--agent conversations are private, costly, or difficult to share. However, useful synthetic dialogues must do more than sound fluent. They should express the intended service goal, maintain coherent interaction flow, and realize the target affective stance in context. We refer to this objective as \emph{pragmatic constraint satisfaction}: the generated dialogue should satisfy specified communicative constraints, including intent, emotion, coherence, and fluency.

We introduce \textsc{PragAlign}, a feedback-guided generate--evaluate--revise framework for controlled synthetic service-dialogue generation. Each dialogue specification contains a service-domain scenario, target intent set, target emotion set, and an auxiliary trait-style control vector. PragAlign first generates a candidate multi-turn dialogue from this specification. An LLM-based evaluator then scores the dialogue for intent alignment, emotion alignment, coherence, fluency, and aggregate quality. If any criterion falls below threshold, the evaluator returns criterion-specific feedback, and the generator revises the dialogue for up to three refinement rounds.

A central challenge is that the evaluator is also part of the refinement loop. We therefore treat automatic acceptance as evaluator-defined constraint satisfaction, not as an independent measure of human-perceived dialogue quality. This distinction is
important because a generator may learn to satisfy the evaluator's rubric without necessarily improving all aspects of human judgment. We therefore organize the evaluation around three questions: whether structured evaluator feedback improves controlled dialogue generation beyond repeated sampling, which communicative constraints benefit most from refinement, and whether evaluator-defined gains correspond to human judgments of intent expression, dialogue flow, and emotion appropriateness. The central empirical question is not whether repeated LLM generation improves outputs, but whether criterion-specific feedback repairs the remaining failures that repeated sampling alone does not reliably fix.

On 800 matched dialogue specifications, PragAlign achieves 99.50\% automatic acceptance, compared with 72.25\% for one-shot generation and 95.88\% for repeated generation without structured feedback. Thus, repeated attempts account for much of the gain over one-shot generation, while structured feedback provides an additional improvement in complete threshold satisfaction. Refinement gains are concentrated in emotion alignment, which is also the dominant failure mode in ablations. A separate human evaluation of 1,200 generated dialogues shows that intent expression and dialogue flow are highly recognizable, while emotion appropriateness is less stable and more subjective.

This paper makes four contributions:
\begin{itemize}
    \item We introduce PRAGALIGN, a closed-loop framework for controlled synthetic service-dialogue generation under explicit intent, emotion, coherence, and fluency constraints.
    \item We evaluate PragAlign on 800 matched dialogue specifications across one-shot, no-feedback, no-emotion, no-personality, and full refinement conditions.
    \item We show that structured evaluator feedback improves complete constraint satisfaction beyond repeated generation alone, with the strongest refinement effects occurring in emotion alignment.
    \item We provide complementary analyses, including ablations and a separate human evaluation, while explicitly treating automatic acceptance as evaluator-defined rather than independent human-quality evidence.
\end{itemize}

\section{Related Work}
\label{sec:related-work}

\paragraph{Iterative refinement and model-based evaluation.}
PragAlign builds on recent work showing that language-model outputs can be improved through feedback, reflection, or deliberative search. Self-Refine and Reflexion use iterative feedback or verbal reflection to improve outputs across multiple attempts, while Tree-of-Thought prompting explores alternative reasoning paths during inference \cite{madaan2023self,shinn2023reflexion,yao2023tree}. In parallel, model-based and LLM-based evaluation methods such as G-Eval, UniEval, and LLM-as-judge frameworks assess generated text quality, task success, or model preferences using structured evaluation criteria \cite{liu2023g,zhong2022towards,zheng2023judging}. PragAlign combines these directions by using an LLM evaluator not only to score generated service dialogues, but also to provide criterion-specific feedback for revision.

\paragraph{Attribute-conditioned dialogue generation.}
The framework is also related to emotion-aware, persona-grounded, user-centric, and personality-conditioned generation. Prior work has explored emotional response generation, empathetic response modeling, persona and knowledge grounding, personality-trait expression in LLM outputs, personality-based synthetic dialogue generation, and controllable multi-attribute dialogue generation \cite{zhou2018emotional,lin2019moel,majumder2020mime,jang2022call,lim2023you,zerhoudi2024personarag,jiang2023personallm,han2024psydial,hu2022controllable}. These approaches show that generated responses can be conditioned on affective, persona, user, personality, or stylistic attributes. PragAlign differs by combining service context, target intents, target emotions, and personality-trait inputs within a closed-loop acceptance procedure for synthetic service-dialogue generation.

\paragraph{Positioning of PragAlign.}
Rather than optimizing for general text quality alone, PragAlign formulates synthetic service-dialogue generation as a communicative-goal alignment problem. Each generated dialogue must satisfy specified constraints for intent alignment, emotion alignment, coherence, fluency, and aggregate quality. The resulting acceptance decision is therefore best interpreted as evaluator-defined pragmatic constraint satisfaction, not as an independent measure of human-perceived dialogue quality.

\section{Model Framework}
\label{sec:model-framework}

\begin{figure}[t]
    \centering
    \includegraphics[width=\linewidth]{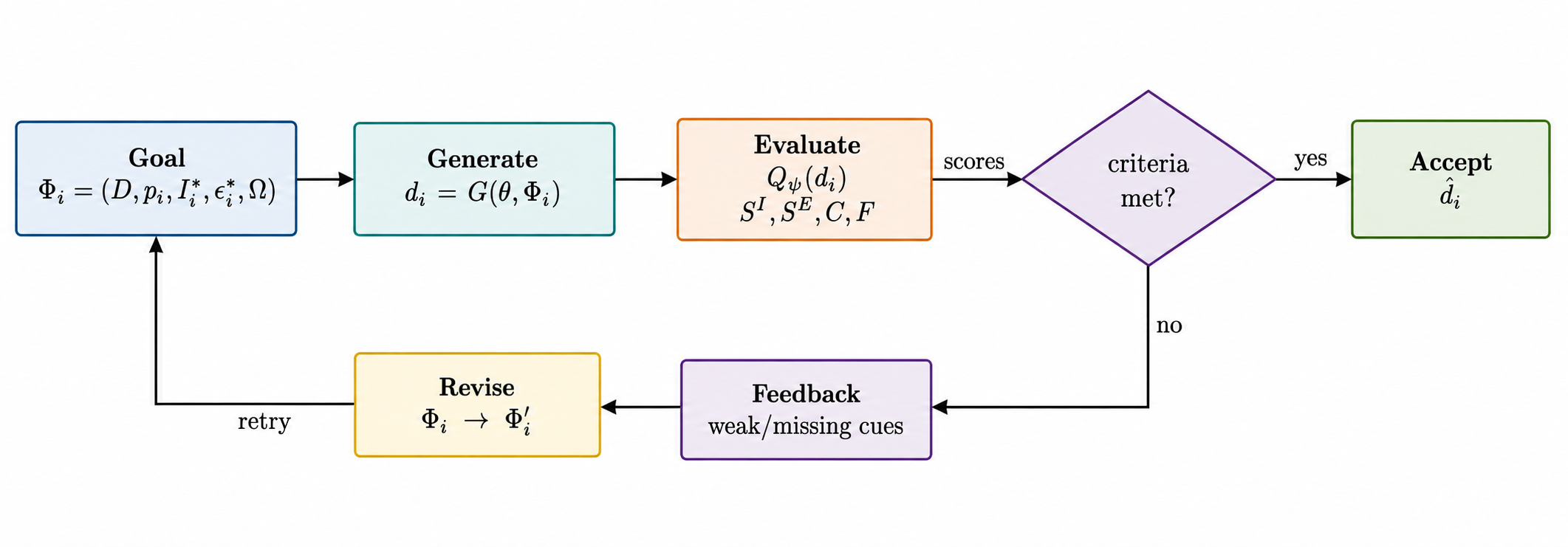}
    \caption{Overview of the PragAlign closed-loop generation framework.}
    \label{fig:PragAlign_overview}
\end{figure}

PragAlign is a goal-driven closed-loop framework for controlled synthetic dialogue generation in customer-facing service interactions. The framework uses an LLM generator to produce candidate multi-turn dialogues and an LLM-based evaluator to assess whether each dialogue satisfies predefined communicative constraints. When a candidate fails one or more criteria, the evaluator returns structured feedback that is incorporated into the next generation attempt. Figure~\ref{fig:PragAlign_overview} summarizes the overall generate--evaluate--revise loop.

\subsection{Goal Representation}
\label{subsec:goal-representation}

Each dialogue is generated from a structured goal representation:
\begin{equation}
\Phi_i = (D, p_i, I_i^{*}, E_i^{*}, \Omega),
\end{equation}
where $D$ denotes the customer-facing service domain context, $p_i$ is an auxiliary trait-style vector used for conditioning, $I_i^{*}$ is the target intent set, $E_i^{*}$ is the target emotion set, and $\Omega$ contains generation hyperparameters. These inputs define the controlled generation space for each candidate dialogue.

The domain specification $D$ grounds generation in a service setting, including the interaction context, user goal, and relevant support details. The target intent set $I_i^{*}$ specifies the functional goals that should be expressed in the dialogue, such as billing clarification, refund request, account access support, complaint handling, or follow-up status request. To construct fine-grained service goals, we use analogical prompting to derive sub-intents from representative examples \cite{yu2023thought}. The target emotion set $E_i^{*}$ specifies affective states based on Plutchik's emotion model \cite{plutchik1980emotion}; these emotions may be expressed directly through emotion words or indirectly through tone, urgency, concern, frustration, relief, or reassurance.

In our implementation, $p_i$ is instantiated using Big Five trait values \cite{de2000big}, but these traits are used only as generation-side style controls and are not treated as evaluated output attributes.

\begin{figure}[t]
    \centering
    \includegraphics[width=\linewidth]{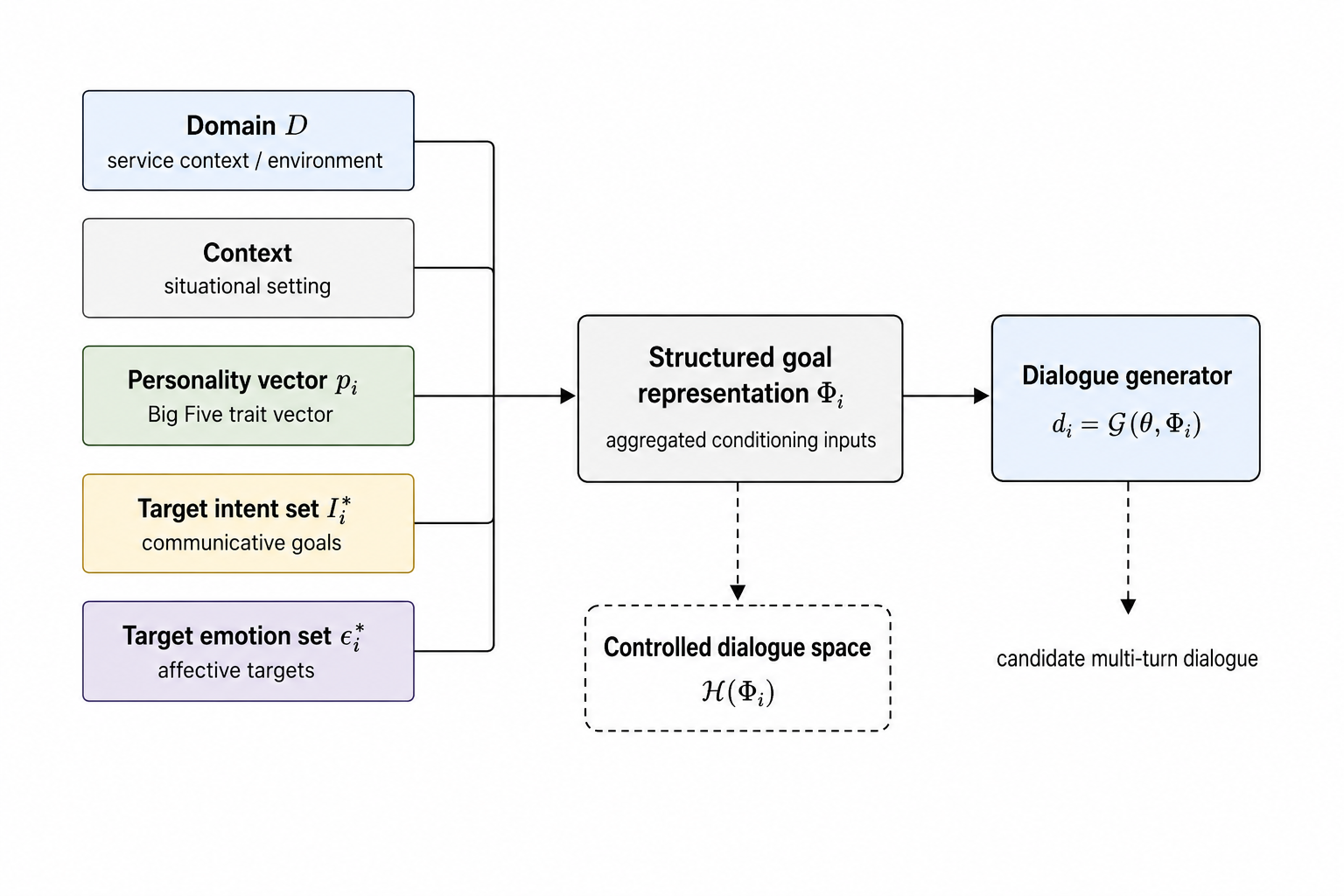}
    \caption{Structured inputs to the dialogue generator.}
    \label{fig:structured_inputs}
\end{figure}

Figure~\ref{fig:structured_inputs} summarizes the structured inputs provided to the dialogue generator.

\subsection{Generate--Evaluate--Revise Loop}
\label{subsec:generate-evaluate-revise}

Given a goal representation $\Phi_i$, the generator produces an initial multi-turn customer-facing service dialogue. Unless otherwise stated, PragAlign uses GPT-4.1 as the dialogue generator. The generation prompt includes the service-domain context, target intents, target emotions, personality traits, and generation constraints such as dialogue length and output format.

After each generation attempt, the evaluator scores the candidate dialogue for intent alignment, emotion alignment, coherence, and fluency. If the dialogue satisfies all acceptance criteria, it is retained as the final output. If any criterion falls below threshold, the evaluator returns criterion-specific feedback identifying the failed dimensions and describing what should be revised. The generator then receives the original goal representation together with this feedback and produces a revised dialogue.

The original communicative specification remains fixed throughout refinement. Thus, revision does not change the target domain, intent set, emotion set, or personality vector; instead, it aims to better realize those constraints in the dialogue. For example, if the evaluator identifies a missing intent or weak emotional realization, the next prompt asks the generator to strengthen that aspect while preserving coherence, fluency, and consistency with the service scenario.

\subsection{LLM-Based Evaluator}
\label{subsec:evaluator}

We implement the evaluator using GPT-4o-mini with structured evaluation prompts. Inspired by criterion-wise decomposition in Tree-of-Thought-style prompting \cite{yao2023tree}, the evaluator assesses each dialogue along separate dimensions for intent alignment, emotion alignment, coherence, and fluency. Each dimension receives a scalar score in $[0,1]$, and the component scores are combined into an aggregate quality score $Q_{\psi}(d_i)$.

In addition to scalar scores, the evaluator produces structured feedback describing weaknesses in the generated dialogue, such as missing intents, weak emotion realization, incoherent turns, or unnatural phrasing. This feedback is used only when a dialogue fails the acceptance criteria; accepted dialogues exit the refinement loop without further revision.

Because the evaluator is also part of the refinement process, automatic acceptance should be interpreted as evaluator-defined constraint satisfaction rather than as independent evidence of human-perceived dialogue quality. The human evaluation is therefore used as a separate dataset-level check of whether generated dialogues contain recognizable cues for intent expression, dialogue flow, and emotion appropriateness.
\section{Experimental Setup}
\label{sec:experimental-setup}

\subsection{Dataset Construction}
\label{subsec:dataset-construction}

We construct 800 matched dialogue specifications for the main automatic evaluation. Each specification contains a service-domain scenario, a target intent set, a target emotion set, and an auxiliary trait-style control vector. The same specifications are used across all automatic experimental conditions to enable paired comparison.

The specifications cover 10 customer-facing service domains: travel booking, online retail delivery, online subscription services, banking support, health appointment services, food delivery support, hotel and hospitality services, utility services, software help desk support, and telecom services. Domain coverage is approximately balanced, with between 72 and 89 specifications per domain.

The intent set contains 15 service intent categories: account access support, billing clarification, service upgrade request, complaint handling, follow-up status request, request invoice or documentation, delivery status inquiry, payment method update, policy clarification, incorrect charge dispute, refund request, confirmation request, replacement or exchange request, contact information update, and escalation request. Each dialogue specification contains either two or three target intents; 150 specifications contain two intents and 650 contain three intents. The emotion set contains 30 affective labels: admiration, amazement, anger, anticipation, apprehension, awe, boredom, contempt, disapproval, disgust, distraction, ecstasy, fear, grief, interest, joy, loathing, love, optimism, pensiveness, rage, remorse, sadness, serenity, submission, surprise, terror, trust, vigilance, and aggressiveness. Each specification contains one to three target emotions; 11 specifications contain one emotion, 118 contain two emotions, and 671 contain three emotions. We also include an auxiliary Big Five trait-style vector for generation-side variation. Each trait takes one of three values, corresponding to low, medium, or high intensity. These values are used only as conditioning inputs and are not evaluated as target output attributes. Generated dialogues were constrained to six to eight turns. Unless otherwise stated, the generator temperature was 0.7, with a maximum output length of 1024 tokens.

\subsection{Implementation Details}
\label{subsec:implementation-details}

Unless otherwise stated, experiments use \texttt{gpt-4.1} as the dialogue generator and \texttt{gpt-4o-mini} as the automatic evaluator. The evaluator is used as an internal quality-control mechanism: it scores generated dialogues, identifies failed criteria, and provides structured feedback for revision. The full PragAlign condition allows up to three refinement rounds after the initial generation attempt.

The no-feedback baseline uses the same maximum number of attempts as full PragAlign, but does not pass criterion-specific evaluator feedback into the next generation prompt. In this condition, each failed attempt is followed by a fresh generation from the original dialogue specification. The generator does not receive the previous dialogue, failed criteria, missing-intent list, missing-emotion list, or evaluator rationale. The maximum number of attempts, generator model, evaluator model, temperature, token limit, and acceptance thresholds are matched to the full PragAlign condition. This baseline isolates the effect of structured feedback from the effect of repeated sampling.

\subsection{Evaluation Protocol}
\label{subsec:evaluation-protocol}

Automatic evaluation measures intent alignment, emotion alignment, coherence, fluency, and aggregate quality. We compute aggregate quality as
\[
Q(d)=\frac{A(d)+C(d)+F(d)}{3},
\]
where $A(d)$ is the mean of intent and emotion alignment, $C(d)$ is coherence, and $F(d)$ is fluency. Because $Q(d)$ can obscure component-level failures, acceptance is determined by both aggregate quality and individual component thresholds. A dialogue is accepted only if it satisfies both the aggregate-quality threshold and all component-level thresholds:

$Q(d) \geq 0.80, S_I(d) \geq 0.80, S_E(d) \geq 0.70, C(d) \geq 0.80, F(d) \geq 0.80$.

We report automatic acceptance as evaluator-defined constraint satisfaction rather than as an independent measure of human-perceived dialogue quality. In addition to the 800-goal automatic evaluation, we conduct a separate human evaluation on 1,200 generated dialogues sampled from the final dataset. Because this human-evaluated set is separate from the matched automatic-evaluation set, we treat it as dataset-level validation rather than as a paired human comparison of experimental conditions.

\section{Results}
\label{sec:results}

We evaluate PragAlign on 800 matched dialogue specifications under five experimental conditions: the full PragAlign framework, one-shot generation, repeated generation without structured feedback, refinement without explicit emotion conditioning, and refinement without personality conditioning. We focus on last-mile constraint satisfaction: cases where generated dialogues are fluent and mostly coherent but still fail one or more required communicative constraints. Each condition uses the same dialogue specifications, enabling paired comparisons across methods. The primary outcomes are automatic acceptance, aggregate quality, component-level alignment scores, refinement behavior, and failure reasons.

We use the term \emph{automatic acceptance} to emphasize that this metric reflects evaluator-defined constraint satisfaction rather than independent human-perceived dialogue quality. The evaluator is part of the refinement loop, so the reported acceptance rate should be interpreted as an internal quality-control measure. A dialogue is accepted only when it satisfies both the aggregate-quality threshold and all component-level thresholds. Thus, a dialogue can receive a high aggregate score while still being rejected if intent alignment, emotion alignment, coherence, or fluency remains below threshold.
\subsection{Overall Automatic Performance}

Table~\ref{tab:overall-auto} reports final automatic performance across the five conditions. Full PragAlign accepted 796 of 800 dialogue specifications, corresponding to an evaluator-defined acceptance rate of 99.50\%. One-shot generation accepted 578 of 800 dialogues, corresponding to 72.25\%. This 27.25 percentage-point difference shows that allowing refinement substantially increases complete threshold satisfaction relative to a single generation attempt.

The more informative comparison, however, is between full PragAlign and the no-feedback repeated-generation baseline, since both conditions allow multiple attempts. The no-feedback baseline accepted 767 dialogues, or 95.88\%, showing that repeated sampling alone accounts for much of the gain over one-shot generation. Full PragAlign accepted 29 additional dialogues, improving acceptance by 3.62 percentage points over no-feedback generation. We therefore interpret structured evaluator feedback as improving \emph{last-mile constraint satisfaction}: it does not dramatically change average dialogue quality, but it helps repair the remaining component-level failures that prevent otherwise strong dialogues from satisfying all acceptance criteria.

This interpretation is supported by the continuous quality scores. Full PragAlign achieved a mean aggregate quality score of 0.953, compared with 0.950 for no-feedback generation and 0.935 for one-shot generation. Thus, the main effect of feedback is not a large shift in average fluency, coherence, or aggregate quality. Instead, feedback increases the probability that a dialogue satisfies every required communicative constraint simultaneously.

Paired analyses further support this distinction. Compared with one-shot generation, full PragAlign produced a mean paired quality improvement of 0.0177, with a bootstrap 95\% confidence interval of [0.0145, 0.0209]. The difference was significant under a Wilcoxon signed-rank test ($p < .001$), with a paired standardized effect size of $d_z = 0.38$. The paired acceptance comparison showed a larger practical effect: full PragAlign accepted 219 specifications that failed under one-shot generation, whereas one-shot generation accepted only one specification that failed under full PragAlign. An exact McNemar test confirmed that the paired acceptance outcomes differed significantly ($p < .001$).

Taken together, these results show that PragAlign's strongest contribution is complete multi-constraint satisfaction. Repeated attempts produce a strong baseline, but structured feedback provides targeted repairs for the remaining cases that fail intent, emotion, coherence, fluency, or aggregate-quality thresholds.

\begin{table*}[t]
\centering
\small
\setlength{\tabcolsep}{4pt}
\begin{tabular}{lrrrrrrrr}
\hline
Condition & Accepted & Acc. (\%) & Quality & Intent & Emotion & Coherence & Fluency & Mean Ref. \\
\hline
Full PragAlign & 796/800 & 99.50 & 0.953 & 1.000 & 0.998 & 0.908 & 0.951 & 0.324 \\
No personality & 798/800 & 99.75 & 0.953 & 1.000 & 0.999 & 0.909 & 0.951 & 0.288 \\
No feedback & 767/800 & 95.88 & 0.950 & 0.999 & 0.985 & 0.907 & 0.952 & 0.456 \\
No emotion & 638/800 & 79.75 & 0.941 & 0.999 & 0.915 & 0.912 & 0.955 & 1.221 \\
One-shot & 578/800 & 72.25 & 0.935 & 0.995 & 0.900 & 0.907 & 0.951 & 0.000 \\
\hline
\end{tabular}
\caption{Automatic performance across five conditions on 800 matched dialogue specifications. Acceptance denotes satisfaction of the configured evaluator-defined thresholds, not independent human judgment.}
\label{tab:overall-auto}
\end{table*}

\subsection{Refinement Dynamics}
\label{subsec:refinement-dynamics}

Table~\ref{tab:refinement-dynamics} summarizes when dialogues were accepted under the full PragAlign condition. Of the 800 dialogues, 590 satisfied all criteria on the initial attempt. The remaining 210 dialogues entered the refinement loop. Among these, 172 were accepted after the first refinement, 27 after the second refinement, and seven after the third refinement. Four dialogues remained unsuccessful after the maximum refinement budget.

\begin{table}[t]
\centering
\small
\begin{tabular}{lrrr}
\hline
Stage & Newly Accepted & Cumulative & Cum. (\%) \\
\hline
Initial generation & 590 & 590 & 73.75 \\
First refinement & 172 & 762 & 95.25 \\
Second refinement & 27 & 789 & 98.63 \\
Third refinement & 7 & 796 & 99.50 \\
Unsuccessful & 4 & -- & 0.50 \\
\hline
\end{tabular}
\caption{Cumulative automatic acceptance across refinement stages for full PragAlign.}
\label{tab:refinement-dynamics}
\end{table}

The largest improvement occurred during the first refinement step. Cumulative acceptance increased from 73.75\% after initial generation to 95.25\% after one refinement. The second refinement added 3.38 percentage points, while the third refinement added 0.87 percentage points. These results show diminishing returns across refinement rounds: most correctable failures were resolved after a single feedback-guided revision.

Raw round-level means should not be interpreted as a standard longitudinal trajectory because dialogues exit the loop once they satisfy the acceptance criteria. Later rounds therefore contain progressively harder unresolved cases. For this reason, we evaluate refinement effectiveness using paired initial-to-final comparisons for the same 210 dialogues that underwent refinement.

\subsection{Initial-to-Final Effects of Refinement}
\label{subsec:initial-final}

Table~\ref{tab:score-improvements} reports paired initial-to-final changes among the 210 full-method dialogues that underwent at least one refinement. Aggregate quality increased by 0.059 on average. Emotion alignment showed the largest component-level improvement, increasing by 0.340 on average. Intent alignment improved by 0.025, while coherence changed only slightly and fluency decreased marginally.

\begin{table}[t]
\centering
\small
\begin{tabular}{lr}
\hline
Metric & Mean Change \\
\hline
Aggregate quality & +0.059 \\
Intent alignment & +0.025 \\
Emotion alignment & +0.340 \\
Coherence & +0.003 \\
Fluency & -0.007 \\
\hline
\end{tabular}
\caption{Mean initial-to-final score changes for refined full PragAlign dialogues.}
\label{tab:score-improvements}
\end{table}

Among the 210 refined dialogues, 204, or 97.14\%, achieved a higher final aggregate score than their initial score. Six dialogues worsened, and none remained exactly unchanged. The paired initial-to-final quality improvement had a large within-dialogue effect size ($d_z = 1.54$). This indicates that refinement reliably improved initially unsuccessful cases under the configured evaluator.

The component-level pattern shows that refinement primarily repaired emotion alignment. Emotion scores improved in 195 refined dialogues and remained unchanged in 15; no refined dialogue exhibited a reduction in emotion alignment. Intent alignment improved in only 15 dialogues and remained unchanged in the remaining 195, suggesting that intent errors were less frequent and more localized. Coherence showed no systematic improvement, and the small decrease in fluency suggests a minor trade-off: revisions targeting missing intent or emotion cues can occasionally introduce slightly less natural phrasing. The magnitude of this fluency decrease was small relative to the normalized score range.

\subsection{Ablation Analysis}
\label{subsec:ablation}

Table~\ref{tab:ablations} compares the full framework with three refinement-based ablations. The results separate three factors: explicit emotion conditioning, structured evaluator feedback, and personality conditioning.

\begin{table*}[t]
\centering
\small
\setlength{\tabcolsep}{4pt}
\begin{tabular}{lrrrrrr}
\hline
Condition & Accepted & Acc. (\%) & Quality & Emotion & Mean Ref. & Failed After Max \\
\hline
Full PragAlign & 796/800 & 99.50 & 0.953 & 0.998 & 0.324 & 4 \\
No emotion & 638/800 & 79.75 & 0.941 & 0.915 & 1.221 & 162 \\
No feedback & 767/800 & 95.88 & 0.950 & 0.985 & 0.456 & 33 \\
No personality & 798/800 & 99.75 & 0.953 & 0.999 & 0.288 & 2 \\
\hline
\end{tabular}
\caption{Ablation comparison on the 800-goal automatic evaluation set.}
\label{tab:ablations}
\end{table*}

Removing explicit emotion conditioning produced the largest degradation. The no-emotion condition accepted 638 of 800 dialogues, corresponding to 79.75\%, compared with 99.50\% for the full framework. It also required substantially more refinement, with a mean of 1.221 refinement rounds compared with 0.324 for full PragAlign. This indicates that explicit emotion conditioning improves both final threshold satisfaction and refinement efficiency.
Removing structured feedback produced a smaller but important degradation. The no-feedback condition accepted 767 dialogues, or 95.88\%, compared with 796 dialogues, or 99.50\%, for full PragAlign. The difference in aggregate quality was small, indicating that repeated generation often produces fluent and coherent dialogues. However, the acceptance gap shows that structured feedback helps with last-mile constraint satisfaction: it repairs specific failed criteria that prevent a dialogue from satisfying all component-level thresholds. Thus, feedback contributes less to broad average-quality improvement than to complete evaluator-defined acceptance.

The no-personality condition accepted 798 dialogues, or 99.75\%, slightly higher than the full framework. Because personality traits are not directly evaluated by either the automatic evaluator or the human protocol, this result should not be interpreted as evidence about personality realization. It shows only that the auxiliary trait-style control does not improve the current intent, emotion, coherence, or fluency metrics.

\subsection{Paired Ablation Statistics}
\label{subsec:paired-ablation}

Table~\ref{tab:paired-ablations} reports paired statistical comparisons between the full method and each ablation. Compared with the no-emotion condition, full PragAlign improved aggregate quality by 0.0113 on average, with a bootstrap 95\% confidence interval of [0.0081, 0.0145]. The Wilcoxon signed-rank test was significant ($p < .001$), and the paired acceptance difference was also significant under McNemar's test ($p < .001$).

Compared with the no-feedback baseline, the mean aggregate-quality difference was small: 0.0024, with a bootstrap 95\% confidence interval of [-0.0002, 0.0051]. The Wilcoxon signed-rank test was not significant ($p = .103$), indicating no reliable difference in continuous aggregate quality. However, the paired acceptance difference was significant under McNemar's test ($p < .001$): full PragAlign accepted 796 dialogues compared with 767 for no-feedback generation. This pattern reinforces the last-mile interpretation. Structured feedback does not substantially raise average quality, but it improves the likelihood that a dialogue satisfies all required thresholds simultaneously.

Compared with the no-personality ablation, full PragAlign showed no measurable advantage. The mean aggregate-quality difference was -0.0003, with a bootstrap 95\% confidence interval of [-0.0028, 0.0023]. The Wilcoxon signed-rank test was not significant ($p = .992$), and the paired acceptance difference was also not significant ($p = .688$).

\begin{table*}[t]
\centering
\small
\setlength{\tabcolsep}{4pt}
\begin{tabular}{lrrrrrrr}
\hline
Comparison & Mean Diff. & 95\% CI & Wilcoxon $p$ & $d_z$ & Full Acc. & Ablation Acc. & McNemar $p$ \\
\hline
Full vs. no emotion & +0.0113 & [0.0081, 0.0145] & $< .001$ & 0.24 & 796 & 638 & $< .001$ \\
Full vs. no feedback & +0.0024 & [-0.0002, 0.0051] & .103 & 0.06 & 796 & 767 & $< .001$ \\
Full vs. no personality & -0.0003 & [-0.0028, 0.0023] & .992 & -0.01 & 796 & 798 & .688 \\
\hline
\end{tabular}
\caption{Paired comparisons between full PragAlign and ablation conditions. Positive quality differences favor the full framework.}
\label{tab:paired-ablations}
\end{table*}

\subsection{Failure Analysis}
\label{subsec:failure-analysis}

Table~\ref{tab:failure-reasons} reports failure reasons by condition. Failure counts are not mutually exclusive because a dialogue can fail more than one criterion. The dominant failure source was emotion misalignment. Under one-shot generation, 214 dialogues failed the emotion-alignment threshold. Under the no-emotion ablation, 162 dialogues failed the same threshold. In contrast, only four full PragAlign dialogues failed due to emotion alignment after refinement.

\begin{table}[t]
\centering
\small
\begin{tabular}{lrrrrr}
\hline
Condition & None & $Q$ & $S_I$ & $S_E$ & $C$ \\
\hline
One-shot & 578 & 5 & 12 & 214 & 0 \\
No feedback & 767 & 0 & 3 & 32 & 0 \\
Full PragAlign & 796 & 0 & 0 & 4 & 0 \\
No emotion & 638 & 5 & 2 & 162 & 0 \\
No personality & 798 & 0 & 1 & 2 & 0 \\
\hline
\end{tabular}
\caption{Failure reasons across conditions. $Q$ denotes aggregate quality below 0.80, $S_I$ intent alignment below 0.80, $S_E$ emotion alignment below 0.70, and $C$ coherence below 0.80. Fluency failures were not observed in this table. Counts are not mutually exclusive.}
\label{tab:failure-reasons}
\end{table}

This pattern is consistent with the paired score analysis. Emotion alignment exhibited the largest initial-to-final improvement and accounted for most of the cases repaired during refinement. At the same time, the persistence of four emotion-related failures after the maximum refinement budget indicates that a small subset of target emotional configurations remained difficult to satisfy even under evaluator-guided revision.

\subsection{Human Evaluation of Generated Dialogues}
\label{subsec:human-eval}

We conducted a separate human evaluation on 1,200 generated dialogues sampled from the final dataset. Because this sample is separate from the 800 matched automatic-evaluation specifications, we treat it as dataset-level validation rather than as a paired human comparison of experimental conditions.

Each dialogue was independently evaluated by two English-speaking annotators at the dialogue level along three binary dimensions: intent expression, dialogue flow, and emotion appropriateness. Annotators were shown the target intent and target emotion, but not the automatic evaluator scores. Personality traits were not evaluated. Table~\ref{tab:human-eval} reports positive rate, raw agreement, and Gwet's AC1 \cite{gwet2008ac1}.

\begin{table}[t]
\centering
\small
\setlength{\tabcolsep}{3pt}
\begin{tabular}{lccc}
\hline
Dimension & Positive Rate & Agreement & AC1 \\
\hline
Intent & 99.7\% & 99.4\% & 0.994 \\
Dialogue Flow & 97.7\% & 95.3\% & 0.951 \\
Emotion & 79.1\% & 75.8\% & 0.639 \\
\hline
\end{tabular}
\caption{Human evaluation results on 1,200 dual-annotated generated dialogues.}
\label{tab:human-eval}
\end{table}

Human annotators judged intent expression and dialogue flow as highly recognizable, while emotion appropriateness was lower and less stable. This suggests that affective realization is more subjective and context-dependent than intent expression or dialogue flow.
\subsection{Automatic--Human Discrepancy for Emotion}
\label{subsec:auto-human-emotion}

The largest automatic--human discrepancy appears in emotion evaluation. Under the automatic evaluator, full PragAlign achieved a mean emotion-alignment score of 0.998, and only four of 800 dialogues failed the automatic emotion threshold. In the separate human evaluation, however, emotion appropriateness received a lower positive rate of 79.1\%, with 75.8\% raw agreement and AC1 of 0.639. This gap suggests that automatic emotion scores should be interpreted cautiously.

One likely reason is that the automatic evaluator may reward explicit affective markers or easily identifiable emotion cues, whereas human annotators may require the target emotion to be contextually appropriate across the full interaction. Emotion cues in service dialogues may also be mild, indirect, or distributed across turns, making them less stable than intent expression or dialogue flow. Thus, the automatic results show that PragAlign satisfies the configured emotion-alignment criterion, while the human results show that reliable human-perceived affective realization remains harder to establish.

\section{Limitations}
\label{sec:limitations}

This study has several limitations. First, the main automatic results depend on an LLM-based evaluator that is also used to guide refinement. The reported 99.50\% automatic acceptance rate should therefore be interpreted as evaluator-defined constraint satisfaction rather than as an independent measure of human-perceived dialogue quality. Because the default generator and evaluator are both from OpenAI model families, future work should reevaluate frozen outputs with independent evaluators from different model families.

Second, the human evaluation is not a paired comparison of experimental conditions. The 1,200 human-evaluated dialogues were sampled separately from the 800 matched automatic-evaluation specifications, so the human study provides dataset-level validation but does not show that humans prefer PragAlign outputs over one-shot or no-feedback outputs for the same goals.

Third, automatic and human emotion results reveal a calibration gap. Full PragAlign reaches near-ceiling automatic emotion alignment, but human annotators rate emotion appropriateness positively in 79.1\% of cases, with lower agreement than for intent expression or dialogue flow. This suggests that the automatic evaluator may reward explicit affective markers without fully capturing contextual appropriateness.

Finally, personality conditioning is not directly validated. PragAlign uses Big Five traits as generation controls, but neither the automatic evaluator nor the human annotation protocol measures whether those traits are recognizable in the generated dialogues. We therefore treat personality as an input conditioning variable rather than a validated output attribute. Additional limitations include ceiling effects in intent, coherence, and fluency metrics, and dependence on hand-selected acceptance thresholds.
\section{Conclusion}
\label{sec:conclusion}

We presented \textsc{PragAlign}, a feedback-guided framework for generating synthetic service dialogues under explicit communicative constraints. Across 800 matched specifications, PragAlign substantially improves evaluator-defined acceptance over one-shot generation and further improves complete constraint satisfaction beyond repeated generation without structured feedback. The results show that closed-loop refinement is especially useful for repairing emotion-alignment failures, which remain the hardest aspect of controlled dialogue generation in both automatic and human evaluation. At the same time, the study highlights the limits of evaluator-guided generation: automatic acceptance is not equivalent to independent human judgment, personality conditioning remains unvalidated, and human annotators find emotion appropriateness less stable than intent expression or dialogue flow. The key lesson is that feedback-guided refinement can enforce explicit communicative constraints, but reliable affective control requires calibration against human judgment rather than evaluator acceptance alone.
\bibliography{custom}
\bibliographystyle{aaai}
\clearpage
\appendix

\section{Appendix}

\subsection{Prompting and Evaluation Setup}
\label{subsec:supp-prompting}

PragAlign uses structured prompts for three stages: initial dialogue generation, evaluator-guided revision, and automatic evaluation. The initial generation prompt conditions the model on the service context, target intents, target emotions, auxiliary trait-style controls, and dialogue-length constraints. The revision prompt preserves the original dialogue specification but adds evaluator feedback describing failed criteria, missing targets, and revision advice. The no-feedback baseline uses the same original dialogue specification and retry budget, but does not provide failed criteria, missing targets, previous dialogue text, or evaluator rationale to the generator.

The automatic evaluator uses a structured rubric with four component dimensions: intent alignment, emotion alignment, coherence, and fluency. For each dialogue, the evaluator returns scalar scores in $[0,1]$ and criterion-specific feedback. Intent and emotion alignment are computed from detected target coverage, while coherence and fluency are scored directly by the evaluator. Aggregate quality is computed as
\[
Q(d)=\frac{A(d)+C(d)+F(d)}{3},
\]
where $A(d)$ is the mean of intent and emotion alignment, $C(d)$ is coherence, and $F(d)$ is fluency.

A dialogue is accepted only if it satisfies both the aggregate-quality threshold and all component-level thresholds: $Q(d) \geq 0.80, S_I(d) \geq 0.80,S_E(d) \geq 0.70, C(d) \geq 0.80, F(d) \geq 0.80.$

This rule is designed to prevent high fluency or coherence from masking failures in target intent or emotion realization.

\subsection{Dataset and Goal Specification}
\label{subsec:supp-goals}

Each dialogue goal is represented as
\[
\Phi_i = (D, p_i, I_i^*, E_i^*, \Omega),
\]
where $D$ is the service-domain context, $p_i$ is an auxiliary trait-style control vector, $I_i^*$ is the target intent set, $E_i^*$ is the target emotion set, and $\Omega$ contains generation hyperparameters.

The main automatic evaluation uses 800 matched dialogue specifications across customer-facing service domains. The same specifications are used across all experimental conditions to support paired comparison. Each specification contains a service scenario, two or three target intents, one to three target emotions, and an auxiliary Big Five-style control vector. The trait-style vector is used only as a generation-side control and is not evaluated as an output attribute.

Generated dialogues are constrained to six to eight turns. Unless otherwise stated, generation uses temperature $0.7$, a maximum output length of 1024 tokens, GPT-4.1 as the dialogue generator, and GPT-4o-mini as the automatic evaluator. The full PragAlign condition allows up to three refinement rounds after the initial generation attempt.

\subsection{Evaluator Prompt and Output Schema}
\label{subsec:supp-evaluator-schema}

The automatic evaluator receives the generated dialogue together with the original dialogue specification and evaluates the dialogue using separate criterion-wise branches. The compact evaluator instruction is:

\begin{quote}
\small
Generated dialogue: \texttt{\{dialogue\}}

Assess the dialogue using separate criterion-wise branches for intent alignment, emotion alignment, coherence, and fluency. Label the utterances with detected target intents and detected target emotions. Provide scalar scores in $[0,1]$, evidence, weaknesses, and revision advice. Return only the structured evaluation object required by the schema.
\end{quote}

The generator returns a dialogue object containing a list of turns:

\begin{verbatim}
{
  "turns": [
    {"speaker": "Customer", "text": "..."},
    {"speaker": "Agent", "text": "..."}
  ]
}
\end{verbatim}

The evaluator returns utterance-level labels and four criterion-wise evaluation branches:

\begin{verbatim}
{
  "utterance_labels": [
    {
      "turn_index": 0,
      "speaker": "Customer",
      "detected_intents": ["..."],
      "detected_emotions": ["..."]
    }
  ],
  "intent_branch": {
    "score": 0.0,
    "evidence": ["..."],
    "weaknesses": ["..."],
    "revision_advice": "..."
  },
  "emotion_branch": {
    "score": 0.0,
    "evidence": ["..."],
    "weaknesses": ["..."],
    "revision_advice": "..."
  },
  "coherence_branch": {
    "score": 0.0,
    "evidence": ["..."],
    "weaknesses": ["..."],
    "revision_advice": "..."
  },
  "fluency_branch": {
    "score": 0.0,
    "evidence": ["..."],
    "weaknesses": ["..."],
    "revision_advice": "..."
  },
  "overall_feedback": "..."
}
\end{verbatim}

This schema allows PragAlign to separate scalar evaluation from revision guidance. The scalar scores determine whether the dialogue satisfies the acceptance criteria, while the weaknesses and revision advice are used only when a dialogue enters the refinement loop.

\subsection{Human Evaluation Protocol}
\label{subsec:supp-human-protocol}

We conducted a separate human evaluation on 1,200 generated dialogues sampled from the final dataset. This sample is separate from the 800 matched automatic-evaluation specifications, so the human evaluation is treated as dataset-level validation rather than as a paired human comparison of experimental conditions.

Each dialogue was independently evaluated by two English-speaking annotators at the dialogue level. Annotators judged three binary dimensions: intent expression, dialogue flow, and emotion appropriateness. They were shown the target intent and target emotion for each dialogue, but not the automatic evaluator scores or refinement history. Personality traits were not evaluated.

We report positive rate, raw agreement, and Gwet's AC1 in the main paper. Gwet's AC1 is used because intent-expression and dialogue-flow labels are highly skewed toward positive judgments, a setting in which Cohen's $\kappa$ can underestimate agreement despite high observed agreement.
\section{Additional Automatic Results}
\label{sec:supp-additional-results}

\subsection{Threshold Sensitivity and Ceiling Effects}
\label{subsec:supp-threshold-sensitivity}

Because automatic acceptance depends on predefined thresholds, we evaluated whether the main conclusions were stable under alternative threshold configurations. Table~\ref{tab:supp-threshold-sensitivity} reports acceptance rates under the default thresholds, a stricter configuration requiring all component scores and aggregate quality to be at least 0.85, a quality-only setting requiring only $Q \geq 0.80$, and a looser all-component setting requiring all criteria to be at least 0.70.

\begin{table}[t]
\centering
\small
\setlength{\tabcolsep}{2.5pt}
\begin{tabular}{lrrrr}
\hline
Condition & Default & Strict & $Q$-only & Loose \\
\hline
One-shot       & 72.25 & 69.63 & 99.38  & 72.25 \\
No feedback    & 95.88 & 91.63 & 100.00 & 95.88 \\
Full PragAlign & 99.50 & 96.25 & 100.00 & 99.50 \\
No emotion     & 79.75 & 77.63 & 99.38  & 79.75 \\
No personality & 99.75 & 95.13 & 100.00 & 99.75 \\
\hline
\end{tabular}
\caption{Acceptance rates under alternative threshold configurations. 
Default uses $Q \geq .80$, $S_I \geq .80$, $S_E \geq .70$, 
$C \geq .80$, and $F \geq .80$. Strict requires all criteria 
to be at least $.85$; $Q$-only requires only $Q \geq .80$; 
Loose requires all criteria to be at least $.70$.}
\label{tab:supp-threshold-sensitivity}
\end{table}

The sensitivity analysis shows that full PragAlign remains above one-shot, no-feedback, and no-emotion conditions under the stricter all-criteria threshold. However, the quality-only setting produces near-ceiling acceptance across all methods. This indicates that aggregate quality alone is insufficient for evaluating controlled dialogue generation, since high fluency and coherence can obscure failures in target intent or emotion realization.

\subsection{Generator Transfer}
\label{subsec:supp-generator-transfer}

To test whether the refinement framework can be applied beyond the primary GPT generator, we evaluated PragAlign using Llama as the dialogue generator while retaining GPT-4o-mini as the automatic evaluator. This experiment used 300 dialogue goals and the full refinement loop.

\begin{table*}[t]
\centering
\small
\setlength{\tabcolsep}{3pt}
\begin{tabular}{lrrrrrrr}
\hline
Generator Setting & $N$ & Accepted & Acc.(\%) & Quality & Intent & Emotion & Coh. \\
\hline
Llama + GPT eval. & 300 & 250 & 83.33 & 0.907 & 0.971 & 0.956 & 0.850 \\
\hline
\end{tabular}
\caption{Llama generator transfer evaluation using GPT-4o-mini as the automatic evaluator. Fluency was 0.909. This is a generator-transfer result, not an independent cross-evaluator result.}
\label{tab:supp-llama-transfer}
\end{table*}

The Llama generator accepted 250 of 300 dialogues, corresponding to an automatic acceptance rate of 83.33\%. Mean quality was 0.907, with intent alignment of 0.971, emotion alignment of 0.956, coherence of 0.850, and fluency of 0.909. These results suggest that the refinement loop can be applied with a non-GPT generator, although performance remains model-sensitive and below the main GPT-based condition.

\subsection{Reproducibility Artifacts}
\label{subsec:supp-artifacts}

The experimental pipeline records final dialogue outputs, iteration histories, evaluator scores, acceptance decisions, failed criteria, and summary tables. The main reproducibility artifacts include final-dialogue files, iteration-history files, automatic summary metrics, failure-reason tables, score-improvement summaries, and human-evaluation export files. These artifacts make it possible to inspect accepted outputs, failed cases, refinement trajectories, and condition-level comparisons.

\end{document}